\documentclass[10pt,twocolumn,letterpaper]{article}

\usepackage[pagenumbers]{cvpr} 
\usepackage{graphicx}
\usepackage{booktabs}
\usepackage{amsmath}
\usepackage{amssymb}
\usepackage{array}
\usepackage{multirow}
\usepackage{color}
\usepackage{colortbl}
\usepackage{framed}
\usepackage{bm}
\usepackage{bbm}
\usepackage{xspace}
\usepackage{enumitem}
\usepackage{xparse}
\usepackage{algorithm}
\usepackage{listings}
\usepackage{url}
\usepackage[dvipsnames]{xcolor}
\usepackage{pifont}
\usepackage{mathtools}
\usepackage[accsupp]{axessibility}

\definecolor{cvprblue}{rgb}{0.21,0.49,0.74}
\usepackage[pagebackref,breaklinks,colorlinks,citecolor=cvprblue]{hyperref}

\newcommand{\cmark}{\ding{51}}%
\newcommand{\xmark}{\ding{55}}%

\definecolor{Gray}{gray}{0.9}
\definecolor{blond}{rgb}{0.98, 0.94, 0.75}
\def \ie {\emph{i.e.}}
\def \eg {\emph{e.g.}}

\newcommand{\tit}[1]{\smallbreak\noindent\textbf{#1.}}
\newcommand{\tinytit}[1]{\noindent\textbf{#1.}}

\newcommand{\ours}{Wiki-LLaVA\xspace}

\newcommand\blfootnote[1]{%
  \begingroup
  \renewcommand\thefootnote{}\footnote{#1}%
  \addtocounter{footnote}{-1}%
  \endgroup
}

\title{Wiki-LLaVA:\\Hierarchical Retrieval-Augmented Generation for Multimodal LLMs}

\author{Davide Caffagni$^{1,*}$ \quad Federico Cocchi$^{1,2,*}$ \quad Nicholas Moratelli$^{1,*}$ \quad Sara Sarto$^{1,*}$ \and Marcella Cornia$^1$ \quad Lorenzo Baraldi$^1$ \quad Rita Cucchiara$^{1,3}$ \\ 
$^1$University of Modena and Reggio Emilia, Italy \quad $^2$University of Pisa, Italy \quad $^3$IIT-CNR, Italy\\
{\tt\small $^1$\{name.surname\}@unimore.it} \quad
{\tt\small $^2$\{name.surname\}@phd.unipi.it}
}
\begin{document}
\maketitle

\begin{abstract}
Multimodal LLMs are the natural evolution of LLMs, and enlarge their capabilities so as to work beyond the pure textual modality. As research is being carried out to design novel architectures and vision-and-language adapters, in this paper we concentrate on endowing such models with the capability of answering questions that require external knowledge. Our approach, termed \ours, aims at integrating an external knowledge source of multimodal documents, which is accessed through a hierarchical retrieval pipeline. Relevant passages, using this approach, are retrieved from the external knowledge source and employed as additional context for the LLM, augmenting the effectiveness and precision of generated dialogues. We conduct extensive experiments on datasets tailored for visual question answering with external data and demonstrate the appropriateness of our approach.
\blfootnote{$^*$Equal contribution.}
\end{abstract}    
\section{Introduction}
\label{sec:intro}
Recently, Large Language Models (LLMs) have demonstrated impressive performance in zero-shot textual tasks. Specifically, recent literature has devised models capable of tackling diverse tasks, as instructed by the user~\cite{ouyang2022training,vicuna2023,touvron2023llama}. In this context, the classical approach is that of fine-tuning a model on varied tasks that are described through natural language~\cite{raffel2020exploring,chung2022scaling}, thus empowering the model to assimilate externally provided instructions and facilitating robust generalization across multiple domains. Following these advancements, the computer vision community has started to investigate the extension of such models to vision-and-language contexts, thus generating Multimodal Large Language Models (MLLMs). On this line, the fusion of visual features into LLM backbones through vision-to-language adapters~\cite{liu2023improved,li2023blip,alayrac2022flamingo,zhu2023minigpt} has induced notable performance improvements, enabling extensive generalization to vision-and-language tasks requiring elaborate visual descriptions.

\begin{figure}[t]
    \centering
    \includegraphics[width=0.96\linewidth]{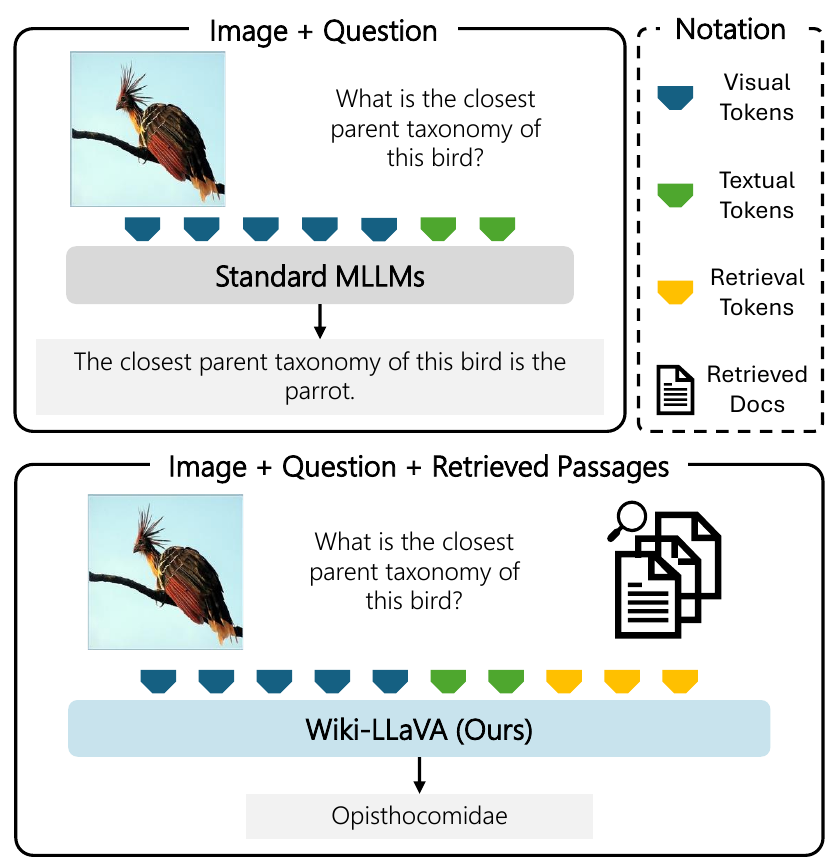}
    \vspace{-0.1cm}
    \caption{Comparison between a standard multimodal LLM and Wiki-LLaVa. Our model integrates knowledge retrieved from an external knowledge base of documents through a hierarchical retrieval pipeline. As a result, it provides more precise answers when tasked with questions that require external knowledge.}
    \label{fig:first_page}
    \vspace{-.3cm}
\end{figure}

In this context, MLLMs excel by simply including a small module (\ie, an adapter) that aligns visual features with textual ones. However, despite these models being built upon LLMs trained on large-scale data, they exhibit notable limitations when confronted with highly specific user queries or when a certain degree of compositional reasoning is required to formulate the response. Moreover, certain knowledge proves itself challenging to be encoded within the parameters of an MLLM, due to the scarcity of long-tail information in the training data. In response to this challenge, different benchmarks have been recently introduced for evaluating the capabilities of MLLM to tackle queries related to external data, such as InfoSeek~\cite{chen2023can} and Encyclopedic-VQA~\cite{mensink2023encyclopedic}. While different  works~\cite{qiu2024snapntell,lerner2024cross,dai2023instructblip,li2023blip} have been testing on these benchmarks, underscoring the significance of this area, none of them has developed architectures specifically designed for tackling external knowledge. 

Driving from these considerations, in this paper we propose the first MLLM augmented with a retrieval module, thus shifting the focus towards teaching the model to leverage diverse information in its responses and learning to discern the relative importance of each. In particular, our model retrieves appropriate information from an external knowledge base of documents and employs a hierarchical retrieval approach to identify relevant passages. This additional knowledge is then fed to an MLLM, without changing its structure but improving its answering capabilities. To the best of our knowledge, our work represents the first MLLM to harness the retrieval capability of external sources. We assess the quality of the proposed approach by conducting extensive experiments and comparisons with respect to recent MLLMs~\cite{liu2023visual,dai2023instructblip,li2023blip} and by showcasing the effectiveness of our design choices. Experimental results demonstrate the advantage of retrieving from external sources and the appropriateness of our model design. Overall, we conceive our work as a first step in the direction of retrieval-augmented MLLMs, which could foster future works in the same area.
\section{Related Work}
\label{sec:related}

\tinytit{Multimodal LLMs}
LLMs have significantly reshaped the landscape of AI research and applications, spearheaded by notable examples like OpenAI's ChatGPT and GPT-4. These models leverage alignment techniques such as instruction tuning~\cite{ouyang2022training} and reinforcement learning from human feedback~\cite{stiennon2020learning} and achieve remarkable capabilities in language understanding and reasoning. Open-source LLMs like Flan-T5~\cite{chung2022scaling}, Vicuna~\cite{vicuna2023}, LLaMA~\cite{touvron2023llama}, and Alpaca~\cite{taori2023stanford} have further accelerated the advancement within the research community. This surge in the development of LLMs subsequently led to the emergence of MLLMs~\cite{caffagni2024r}, which can combine the understating of visual inputs with natural language generation.

Early attempts of building MLLMs such as VisualGPT~\cite{chen2022visualgpt} and Frozen~\cite{tsimpoukelli2021multimodal} used pre-trained language models to enhance vision-and-language models specifically for tasks like image captioning and visual question answering. This initial investigation paved the way for subsequent research in this domain, with the introduction of solutions such as Flamingo~\cite{alayrac2022flamingo} or BLIP-2~\cite{li2023blip} which allowed the integration of image features into LLMs respectively through trainable cross-attention layers directly within the LLM or Q-Former blocks that instead combine image and textual features via learnable queries. Building upon these advancements, subsequent models like FROMAGe~\cite{koh2023grounding}, Kosmos-1~\cite{huang2023language}, and MiniGPT-4~\cite{zhu2023minigpt} have been introduced to further refine the interplay between visual and language modalities within the LLM architecture. 

Concurrently, the LLaVA family of models~\cite{liu2023visual,liu2023improved,liu2024llavanext} introduced the usage of instruction tuning in the multimodal domain, by training on a curated dataset collected with GPT-4. This strategy is now among the most promising recipes for building MLLMs.

\tit{Retrieval-augmented language models} 
In recent years, retrieval-augmentation has been applied to language models by expanding their input space with relevant text passages extracted from external sources~\cite{guu2020retrieval} or eventually retrieved directly from the web~\cite{nakano2021webgpt}. These techniques have demonstrated large
improvements in knowledge-intensive tasks and significant savings in terms of model size. 

Traditionally, the integration of external knowledge into textual generation has been confined to the initial stages. Different solutions~\cite{jiang2023active} proposed to adaptively retrieve passages for generation on top of a proprietary LLM. Some works~\cite{guu2020retrieval}, instead, focused on capturing knowledge in a more modular and interpretable way, by augmenting the language model pre-training with a latent knowledge retriever. This
allows the model to retrieve and attend documents taken from a large corpus such as Wikipedia.

While much attention has been directed towards textual augmentation, similar research efforts have recently been dedicated in the context of vision-and-language tasks~\cite{barraco2023little,sarto2023positive,hu2023reveal,poppi2024removing}. Following this direction, the work presented in~\cite{hu2023reveal} proposed a retrieval-augmented visual-language model that encodes world knowledge into a large-scale memory. Other approaches~\cite{sarto2022retrieval,ramos2023smallcap} also apply retrieval to specific downstream tasks such as image captioning. Differently from all the aforementioned approaches, our work is the first to apply retrieval-augmentation to MLLMs. We do this by applying a hierarchical retrieval strategy on top of a knowledge base made of multimodal documents.

\begin{figure*}[t]
    \centering
    \includegraphics[width=0.98\textwidth]{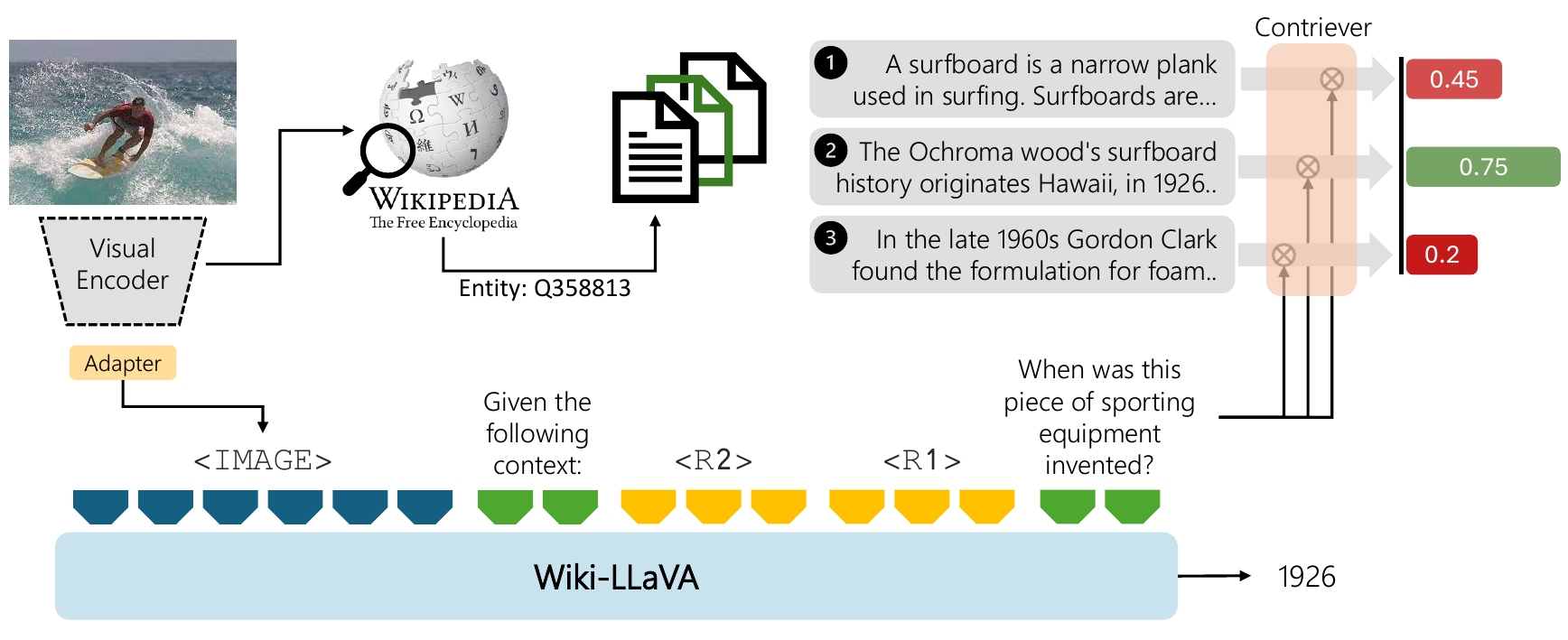}
    \vspace{-.1cm}
    \caption{Overview of the architecture of \ours, which augments a multimodal LLM with external knowledge through a hierarchical retrieval pipeline.}
    \label{fig:model}
    \vspace{-.3cm}
\end{figure*}

\tit{Knowledge-based visual question answering} 
Recently, the emergence of new benchmarks like Encyclopedic-VQA~\cite{mensink2023encyclopedic} and InfoSeek~\cite{chen2023can} has raised the difficulty of standard knowledge-based VQA~\cite{marino2019ok, jain2021select, schwenk2022okvqa} with questions that require intensive knowledge about specific entities, such that even LLM-based models perform poorly without retrieving information from external sources. Often, contrastive image-text encoders are employed to retrieve the target entity given the query image~\cite{wei2023uniir, xiao2024grounding}. Then, the entity name is used as a key to access an external knowledge base, which is typically composed of several text passages that encompass the correct answer. In this work, we design a hierarchical retrieval scheme based on CLIP~\cite{radford2021learning} and the Contriever model~\cite{izacard2021unsupervised} to extrapolate relevant passages, and we feed them to an MLLM to help the answer generation.

\section{Proposed Method}
\label{sec:method}
Our goal is to equip Multimodal LLMs (MLLMs) with the ability to answer complex and specific questions that cannot be addressed solely through the image content and pre-trained knowledge. To achieve this, we propose \ours, which integrates external knowledge derived from an external memory into the LLaVA model, without significantly altering its design. Instead, we augment the capabilities of the model by incorporating retrieval information as additional input context. 
Overall, \ours comprises three components, as shown in Fig.~\ref{fig:model}: a visual encoder, which is employed to provide the MLLM with visual context and as a query to retrieve from an external knowledge base, the knowledge base itself (\eg, Wikipedia), and a hierarchical retrieval module which retrieves relevant documents and passages from the external knowledge base, to be employed as additional context for the MLLM.

\subsection{Knowledge-based Augmentation}
\tinytit{Multimodal integration and autoregressive generation}
An MLLM usually takes as input a multimodal input query, comprising both image and text, and generates a textual output in an autoregressive manner. Formally, the architecture is trained to model a probability distribution $p(w_t|I, w_0, w_1, ..., w_{t-1}, \theta)$, where $\theta$ denotes the parameters of the model, $I$ represents an input image, and ${w_0,..,w_{t-1}}$ denotes the textual prompt. The textual prompt usually includes a pre-defined system-level prompt and a question related to the input image, given by the user. Clearly, a standard MLLM can only rely on the user prompt, the input image, and the knowledge stored in its internal parameters (\ie, $\theta$) to accommodate requests, thus limiting its ability to answer questions that rely on external knowledge. 

In the rest of the paper, we employ LLaVA~\cite{liu2023visual} as our reference MLLM. LLaVA exploits the capabilities of a pre-trained LLM (\ie, Vicuna~\cite{vicuna2023}) and a pre-trained visual model (\ie, a CLIP-based visual encoder~\cite{radford2021learning}), which are interconnected through an MLP adapter, in charge of converting CLIP features to dense input tokens. For an input image $I$, therefore, LLaVA utilizes a pre-trained CLIP visual encoder $E_v$, extracts a dense grid of visual features $Z_v = E_v(I)$, which is then projected via a learnable MLP to produce a sequence of dense embedding tokens $v_o, v_1, ..., v_N$. Finally, these are prepended to the system prompt, and the full sequence of visual and textual tokens is then given as input to the LLM component of the model.

\tit{Augmentation with external knowledge}
To augment the MLLM with external knowledge, we enrich the input context by injecting relevant textual data from an external memory composed of documents. Formally, the distribution of the MLLM is conditioned on additional textual retrieval-knowledge tokens, leading to
\begin{equation}
    p(w_t|
    \overbracket{v_o, v_1, ..., v_N}^{\text{Visual tokens}},\ \ \ \ \underbracket{w_0, w_1, ..., w_{t-1}}_\text{\textcolor{red}{System + user prompt}}, \overbracket{e_0, e_1, ..., e_{\tau}}^\text{\textcolor{blue}{External memory tokens}}),    
\end{equation}
where $e_0, ..., e_{\tau}$ represents the added tokens retrieved from the external memory. Differently from the standard formulation of MLLMs, by enriching the input context we allow the model to generate more specific answers by exploiting tokens retrieved from the memory.

\tit{Hierarchical retrieval from an external memory}
The external memory comprises a collection of (document, image, text-title) triplets taken from documents, denoted as $\mathcal{D} = \{(d_i, t_i)_i\}$. Within this memory, we conduct a hierarchical two-step search to retrieve appropriate information. Initially, we locate the most pertinent document, followed by identifying the relevant passage inside a particular document, which is subsequently exploited as additional input context in the MLLM.

In the first stage, given an input query image $I$ we perform an approximate $k$-nearest neighbor search into the external memory, using document titles as retrievable keys. The similarity between the query image and the text titles is modeled as the inner product between their respective embeddings, which are computed through the visual and textual CLIP encoders (\ie, $E_v$ and $E_t$), as follows:
\begin{equation}
    \text{sim}(I_i, t_i) = E_v(I) \cdot E_t(t_i)^T.
\end{equation}
Then, the knowledge retriever returns the top-$k$ documents associated with the most relevant items retrieved using the aforementioned procedure.

\tit{Retrieving document passages}
In the second step, we analyze each of the retrieved documents to identify the most relevant passages corresponding to the user's question. 
Each document is defined as a sequence of chunks, denoted as $d_i = [c_{i_0},..,c_{i_T}]$, and, given the input question, we retrieve the chunks with the highest similarity to the question. We employ the Contriever architecture~\cite{izacard2021unsupervised} to embed each chunk of the selected document, along with the query (\ie, the question provided by the user), and compute the similarity as an inner product between embeddings. By retrieving the $n$ most appropriate passages inside each of the retrieved documents, overall we obtain $k \cdot n$ passages.

\tit{Context enrichment}
Once we find the most relevant chunks, we employ their raw contents as an additional input to the MLLM. Specifically, the final prompt that we employ includes the image tokens, the retrieved raw chunks, the system-level prompt, and the user question. Formally, considering three retrieved passages, the final prompt is defined as follows:
\begin{gather}
  \small
   \nonumber
   \texttt{<IMAGE>\textbackslash nGiven the following context:\textbackslash n} \\
   \nonumber
   \small
    \texttt{ <R1>\textbackslash n<R2>\textbackslash <R3>\textbackslash n <QUESTION>} \\
    \small
    \texttt{Give a short answer. ASSISTANT:}
\end{gather} 

\subsection{Training}
\label{sec:training}
While the aforementioned approach could work in a zero-shot fashion, using the original weights $\theta$ of the pre-trained MLLM, we also investigate the case of fine-tuning the model to augment its capabilities of exploiting retrieved passages. In particular, in this case, the model is trained on pairs of questions and ground-truth answers requiring external knowledge. As this would potentially reduce the capabilities of the MLLM on tasks not requiring external knowledge (\ie, all the other tasks on which the model has been originally trained), we apply a data mixing approach in which ground-truth pairs requiring external knowledge are mixed with ground-truth pairs not requiring external knowledge in the same mini-batch.

\section{Experiments}
\label{sec:experiments}
In this section, we first introduce the experimental settings, describing the datasets employed, the evaluation protocol, and the implementation and training details used to perform the experiments. Then, we present our experimental results, analyzing the effectiveness of CLIP fine-tuning and evaluating how it is possible to incorporate retrieved knowledge in an MLLM. Finally, limitations of the proposed approach and possible future works are reported.

\subsection{Datasets}
\tinytit{Encyclopedic-VQA~\cite{mensink2023encyclopedic}} 
The dataset contains around 221k question-answer pairs associated with 16.7k different fine-grained entities, with up to 5 images representing the same entity. Overall, there are more than 1M triplets composed of an image, a question, and the corresponding answer. Fine-grained entities and related images are extracted from iNaturalist 2021~\cite{van2021benchmarking} and Google Landmarks Dataset V2~\cite{weyand2020google}, which are associated with the corresponding Wikipedia article. Questions are divided into four different categories, namely single-hop, automatically generated, multi-answer, and two-hop. In particular, single-hop questions have been manually annotated and a single Wikipedia article is needed to answer them. Automatically generated questions are similar to the single-hop questions but have been generated by automatic models. Multi-answer questions, instead, can be answered with a list of terms, but always refer to a single fine-grained entity. Finally, two-hop questions require two retrieval steps to answer them. The dataset also comes with a knowledge base composed of 2M Wikipedia articles, suitable for answering dataset questions.

Dataset triplets are divided into training, validation, and test splits respectively composed of 1M, 13.6k, and 5.8k samples. In our experiments, we employ the training split to fine-tune the LLaVA model and report the results on the test set of the dataset. During testing, we filter out two-hop questions resulting in 4,750 test triplets.

\tit{InfoSeek~\cite{chen2023can}} The dataset contains 1.3M image-question-answer triplets corresponding to around 11k different entities (\ie, Wikipedia articles). The vast majority of questions have been obtained with an almost entirely automatic procedure, by filling human-authored templates with knowledge triples from Wikidata. In this case, images are derived from the OVEN dataset~\cite{hu2023open}. Triplets are divided into training, validation, and test sets, with around 934k, 73k, and 348k samples respectively. At the time of the submission, the ground-truth answers and entities from the test set were not available. Therefore, we report our results on the validation split. Both validation and test sets contain questions related to new entities not included in the training split and questions not seen during training.  

Along with image-question-answer triplets, a knowledge base composed of 6M Wikipedia entities is provided. In our experiments, we consider a randomly extracted subset of 100k entities, in which we guarantee the presence of the 6,741 entities associated with questions from the training and validation splits.

\subsection{Implementation Details}
\label{sec:details}
\tinytit{LLaVA fine-tuning} We employ two distinct fine-tuning approaches, with each being exclusively applied to one of the datasets. In order to maintain the performance of the LLaVA model on well-established MLLM datasets, we supplement fine-tuning data with samples from the LLaVA-Instruct dataset~\cite{liu2023visual}. Specifically, given its size of 158k, we double the probability of having examples from this dataset in each mini-batch. To reduce the number of trainable parameters, we train using low-rank adapters~\cite{hu2021lora} with a total batch size of 512 samples.

\tit{Retrieval} Textual documents sourced from Wikipedia content are embedded using the Contriever architecture~\cite{izacard2021unsupervised}, segmenting the text into chunks of 600 characters each. Furthermore, for streamlined efficiency, the process involves utilizing a single visual encoder. Specifically, following the LLaVA architecture~\cite{liu2023visual}, we employ the CLIP ViT-L/14@336 backbone to embed images to give as input to the MLLM, while simultaneously leveraging it to extract query visual features in the initial hierarchical retrieval step, facilitating the integration of an external memory component. 

To perform entity retrieval, we employ approximate \textit{k}NN search rather than exact \textit{k}NN search because it significantly improves
the computational speed of the entire pipeline. To this aim, we employ the Faiss library~\cite{johnson2019billion} and a graph-based HNSW index with 32 links per vertex.

\subsection{Evaluation Protocol} 
We evaluate our models in two settings: without external knowledge base and with external knowledge base. The former means that we ask the model to directly answer a visual question, by solely relying on the competencies learned during pre-training and/or fine-tuning. On the other hand, in the latter setting, we leverage the proposed hierarchical retrieval method to search for additional information in the external knowledge base. In practice, this is represented by two dumps of Wikipedia comprehending 2M and 100k pages, respectively for Encyclopedic-VQA and InfoSeek. Concerning the evaluation metrics, we report the accuracy over the Encyclopedic-VQA test split and the InfoSeek validation split, following the official evaluation scripts provided along with the datasets.

\subsection{Experimental Results}
\tinytit{Analyzing CLIP performance} We start by evaluating entity retrieval results using CLIP. In this setting, we consider images from the Encyclopedic-VQA test set and InfoSeek validation set and measure the CLIP ability to find the correct entity within the knowledge base of each respective dataset (\ie, composed of 2M entries for Encyclopedic-VQA and 100k entries for InfoSeek). As previously mentioned, we perform retrieval using images as queries and Wikipedia titles as retrievable items.

Results are reported in Table~\ref{tab:retrieval} in terms of recall@$k$ (R@$k$) with $k=1,10,20,50$ which measures the percentage of times the correct entity is found in the top-$k$ retrieved elements. Notably, correctly retrieving the Wikipedia entity associated with the input image strongly depends on the size of the employed knowledge base. In fact, when using 100k items, as in the case of InfoSeek, the correct entity is retrieved as the first item 36.9\% of the time and among the top-10 66.1\% of the time. Instead, when using a significantly larger knowledge base as in the case of Encyclopedic-VQA, which contains 2M items, retrieval results are significantly lower with 3.3\% and 9.9\% respectively in terms of R@1 and R@10.

\begin{table}[t]
  \centering
  \setlength{\tabcolsep}{.45em}
  \resizebox{0.95\linewidth}{!}{
  \begin{tabular}{lcc cccc}
   \toprule
    \textbf{Dataset} & \textbf{KB} & & R@1 & R@10 & R@20 & R@50 \\
    \midrule
    Encyclopedic-VQA & 2M & & 3.3 & 9.9 & 13.2 & 17.5 \\
    InfoSeek & 100k & & 36.9 & 66.1 & 71.9 & 78.4 \\
  \bottomrule
  \end{tabular}
  }
  \vspace{-.1cm}
  \caption{Entity retrieval results on the Encyclopedic-VQA test set and InfoSeek validation set. To comply with the visual encoder employed in LLaVA, all results are obtained using CLIP ViT-L/14@336.}
  \label{tab:retrieval}
  \vspace{-0.2cm}
\end{table}

\begin{table*}[t]
  \centering
  \setlength{\tabcolsep}{.45em}
   \resizebox{0.76\linewidth}{!}{
  \begin{tabular}{lcccccc cc c ccc}
   \toprule
    & & & & & & & \multicolumn{2}{c}{\textbf{Enc-VQA}} & & \multicolumn{3}{c}{\textbf{InfoSeek}} \\
    \cmidrule{8-9} \cmidrule{11-13} 
    \textbf{Model} & \textbf{LLM} & & \textbf{KB} & $k$ & $n$ & & Single-Hop & All & & Unseen-Q & Unseen-E & All \\
    \midrule
    \textbf{Zero-shot Models} \\
    \hspace{0.4cm}BLIP-2~\cite{li2023blip} & Flan-T5$_\text{XL}$ & & \xmark & - & - & & 12.6 & 12.4 & & 12.7 & 12.3 & 12.5 \\
    \hspace{0.4cm}InstructBLIP~\cite{dai2023instructblip} & Flan-T5$_\text{XL}$ & & \xmark & - & - & & 11.9 & 12.0 & & 8.9 & 7.4 & 8.1 \\
    \hspace{0.4cm}LLaVA-1.5~\cite{liu2023improved} & Vicuna-7B & & \xmark & - & -  & & 16.3 & 16.9 & & 9.6 & 9.4 & 9.5 \\
    \midrule
    \textbf{Fine-tuned Models} \\
    \hspace{0.4cm}LLaVA-1.5~\cite{liu2023improved} & Vicuna-7B & & \xmark & - & -  & & 23.3 & 28.5 & & 19.4 & 16.7 & 17.9 \\
    \rowcolor{blond}
    \hspace{0.4cm}\textbf{\ours} & Vicuna-7B & & \cmark & 1 & 1 & & 21.8 & 26.4 & & 26.6 & 24.6 & 25.5 \\
    \midrule
    \rowcolor{blond}
    \hspace{0.4cm}\textbf{\ours} & Vicuna-7B & & \cmark & 1 & 2 & & 19.9 & 23.2 & & 29.1 & 26.3 & 27.6\\
    \rowcolor{blond}
    \hspace{0.4cm}\textbf{\ours} & Vicuna-7B & & \cmark & 1 & 3 & & 17.7 & 20.3 & & 30.1 & 27.8 & 28.9 \\
    \midrule
    \rowcolor{blond}
    \hspace{0.4cm}\textbf{\ours} & Vicuna-7B & & \cmark & 2 & 1 & & 21.3 & 25.4 & & 27.8 & 24.6 & 26.1 \\
    \rowcolor{blond}
    \hspace{0.4cm}\textbf{\ours} & Vicuna-7B & & \cmark & 3 & 1 & & 20.5 & 24.3 & & 27.4 & 24.5 & 25.3 \\
    \midrule
    \midrule
    \rowcolor{Gray}
    \hspace{0.4cm}\textbf{\ours} & Vicuna-7B & & \cmark & 1 & 1 & & 34.7 & 37.2 & & 41.1 & 41.1 & 41.1 \\
    \rowcolor{Gray}
    \hspace{0.4cm}\textbf{\ours} & Vicuna-7B & & \cmark & 1 & 2 & & 39.2 & 40.2 & & 49.1 & 46.5 & 47.8 \\
    \rowcolor{Gray}
    \hspace{0.4cm}\textbf{\ours} & Vicuna-7B & & \cmark & 1 & 3 & & 38.5 & 38.6 & & 52.7 & 50.3 & 51.5 \\
  \bottomrule
  \end{tabular}
  }
  \vspace{-.1cm}
  \caption{Accuracy results on the Encyclopedic-VQA test set and InfoSeek validation set. \colorbox{blond}{\textbf{Yellow color}} indicates models employing the CLIP model to perform entity retrieval, while \colorbox{Gray}{\textbf{gray color}} indicates the use of ground-truth entities (\ie, oracle). $k$ denotes the number of retrieved entities, and $n$ represents the number of textual chunks retrieved for each entity that are given to the MLLM as additional context.}
  \label{tab:results}
\vspace{-.3cm}
\end{table*}

\tit{Results on Encyclopedic-VQA and InfoSeek}
We then report visual question-answering results in Table~\ref{tab:results}. We include the performance of zero-shot models like BLIP-2~\cite{li2023blip}, InstructBLIP~\cite{dai2023instructblip}, and the LLaVA-1.5 baseline model~\cite{liu2023visual}, which are not fine-tuned on the considered datasets and that do not leverage the external knowledge base. Moreover, we consider the accuracy results of LLaVA-1.5 when fine-tuned on the training set of Encyclopedic-VQA and InfoSeek, but not augmented with retrieved context. The results of our approach (\ie, \ours) are reported both in the standard setting in which CLIP is used to retrieve the most representative entity from the knowledge base and in its \textit{oracle} version, which employs the entity corresponding to the input image-question pair. For both cases, we consider a different number $n$ of retrieved textual chunks, all corresponding to the top-1 (or ground-truth) entity. When employing CLIP, we also vary the number $k$ of retrieved entities (\ie, $k=1,2,3$) using $n=1$ when $k$ is greater than 1. This choice is given by the maximum context length that Vicuna takes as input, which is set to 2,048 tokens.

As it can be seen, zero-shot MLLMs face difficulties in correctly answering the given questions as these models can only rely on the knowledge embedded inside the LLM. When instead using an external knowledge base, the accuracy results significantly increase especially on the InfoSeek dataset with 100k retrievable items. The limited performance of the CLIP model in retrieving the correct entity on larger knowledge bases, instead, leads to a slight degradation of accuracy scores. This is due to the noisy textual passages that are provided to the MLLM as additional external context which, being related to a different entity, often do not contain informative content.

Overall, retrieving passages from different entities does not always help increase the results. Instead, using more than one textual chunk as additional context for the MLLM generally improves the final accuracy on the InfoSeek validation set with an overall improvement of 2.1 and 3.4 accuracy points with $n=2$ and $n=3$ respectively. Furthermore, it is worth noting that employing oracle entities significantly boosts the final accuracy. In particular, oracle entities lead to an improvement of 13.8\% on Encyclopedic-VQA and 22.6\% on InfoSeek, comparing the best-performing configuration with CLIP-based entity retrieval (\ie, $k=1$ and $n=1$ for Encyclopedic-VQA and $k=1$ and $n=3$ for InfoSeek) with the best performing oracle-based version (\ie, $k=1$ and $n=2$ for Encyclopedic-VQA and $k=1$ and $n=3$ for InfoSeek). These results confirm the effectiveness of directly employing retrieved passages to augment a pre-trained MLLM and further highlight the importance of having a good entity retrieval model to limit the possibility of feeding the MLLM with irrelevant content.

\begin{table}[t]
    \centering
    \setlength{\tabcolsep}{.3em}
    \resizebox{\linewidth}{!}{
    \begin{tabular}{lc cc c ccc}
    \toprule
         & & \multicolumn{2}{c}{\textbf{Enc-VQA}} & & \multicolumn{3}{c}{\textbf{InfoSeek}} \\
         \cmidrule{3-4} \cmidrule{6-8} 
         \textbf{Fine-tuning} & & Single-Hop & All & & Unseen-Q & Unseen-E & All \\
         \midrule
         \xmark & & 16.3 & 16.9  & & 9.6 & 9.4 & 9.5 \\
         \midrule
         \cmark & & 23.4 & 29.0 & & 17.1 & 15.0 & 16.0 \\
         \cmark~+ LLaVA-Instruct & & 23.3 & 28.5 & & 19.4 & 16.7 & 17.9 \\
    \bottomrule
    \end{tabular}
    }
    \vspace{-.1cm}
    \caption{Performance analysis when using the LLaVA-Instruct dataset during fine-tuning. All results are obtained without external knowledge retrieval.}
    \label{tab:ablation}
      \vspace{-0.35cm}
\end{table}

\begin{figure*}[t]
\centering
    \begin{minipage}{0.165\linewidth}
        \includegraphics[width=0.97\linewidth]{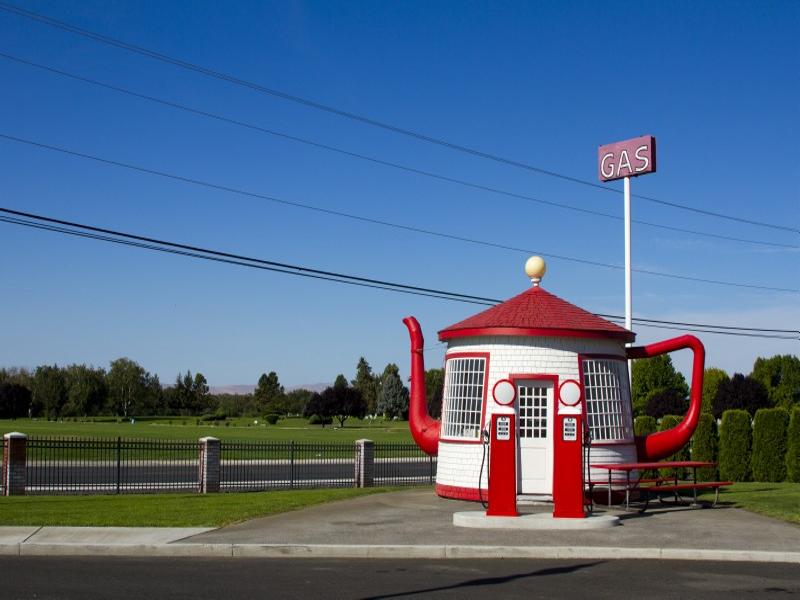}
        \end{minipage}
        \begin{minipage}{0.16\linewidth}
        \scriptsize{
	In what state is this building located? \vspace{0.15cm}\\
        \textbf{LLaVA-1.5:}\\
        California \textcolor{red}{\xmark} \\
        \textbf{\ours:}\\
        Washington \textcolor[HTML]{00b050}{\cmark}
        }
    \end{minipage}
    \hspace{0.02cm}
        \begin{minipage}{0.165\linewidth}
        \includegraphics[width=0.97\linewidth]{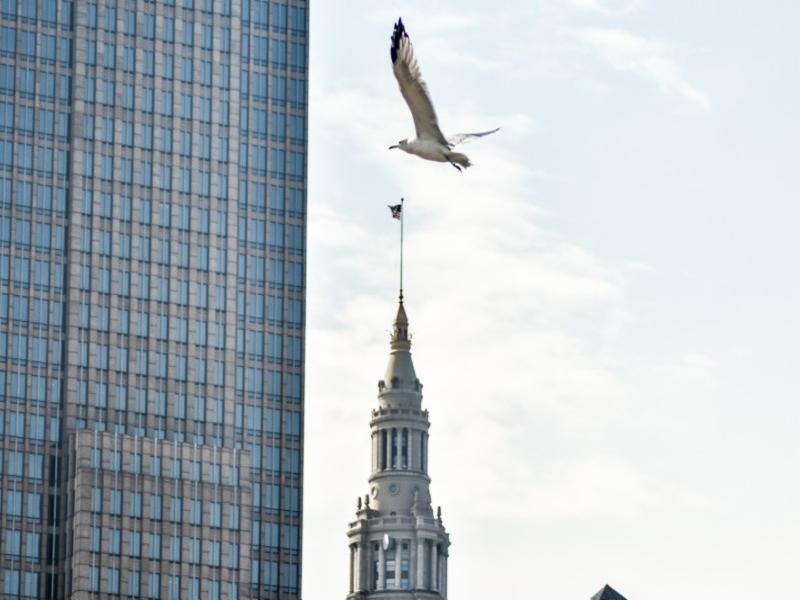}
        \end{minipage}
        \begin{minipage}{0.15\linewidth}
        \scriptsize{
	When was this building constructed? \vspace{0.15cm}\\
        \textbf{LLaVA-1.5:}\\
        1970 \textcolor{red}{\xmark} \\
        \textbf{\ours:}\\
        1927 \textcolor[HTML]{00b050}{\cmark}
        
        }
    \end{minipage}
    \hspace{0.02cm}
        \begin{minipage}{0.165\linewidth}
        \includegraphics[width=0.97\linewidth]{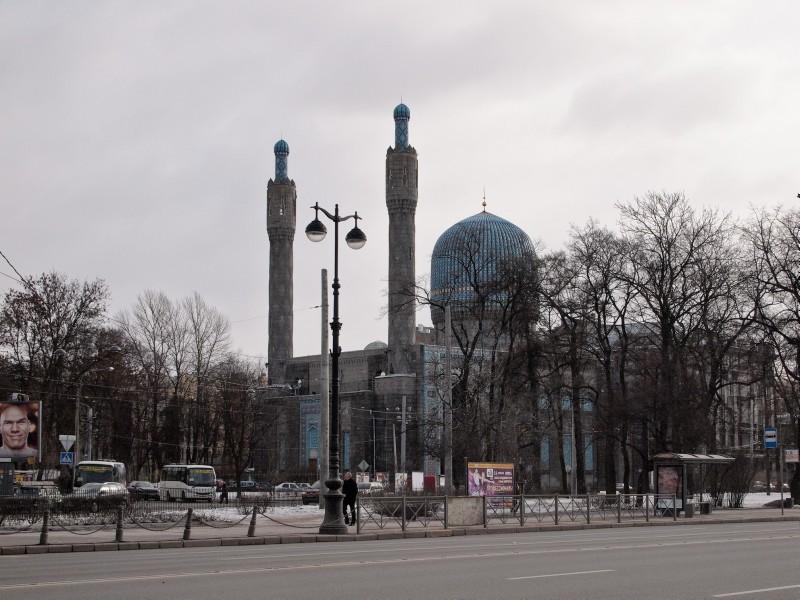}
        \end{minipage}
        \begin{minipage}{0.15\linewidth}
        \scriptsize{
	What's the height of the tallest minaret from this mosque? \vspace{0.15cm}\\
        \textbf{LLaVA-1.5:}\\
        100 feet \textcolor{red}{\xmark} \\
        \textbf{\ours:}\\
        49mt \textcolor[HTML]{00b050}{\cmark}
        }
    \end{minipage}
    
    \vspace{0.15cm}

    \begin{minipage}{0.165\linewidth}
        \includegraphics[width=0.97\linewidth]{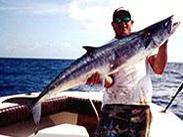}
        \end{minipage}
        \begin{minipage}{0.16\linewidth}
        \scriptsize{
	Which geographic area is this fish found? \vspace{0.15cm}\\
        \textbf{LLaVA-1.5:}\\
        Gulf of Mexico \textcolor{red}{\xmark} \\
        \textbf{\ours:}\\
        Brazil \textcolor[HTML]{00b050}{\cmark}
        }
    \end{minipage}
    \hspace{0.02cm}
        \begin{minipage}{0.165\linewidth}
        \includegraphics[width=0.97\linewidth]{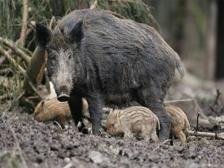}
        \end{minipage}
        \begin{minipage}{0.15\linewidth}
        \scriptsize{
	What is the oldest age of this animal? \vspace{0.15cm}\\
        \textbf{LLaVA-1.5:}\\
        10 years \textcolor{red}{\xmark} \\
        \textbf{\ours:}\\
        24.9 \textcolor[HTML]{00b050}{\cmark}
        }
    \end{minipage}
    \hspace{0.02cm}
        \begin{minipage}{0.165\linewidth}
        \includegraphics[width=0.97\linewidth]{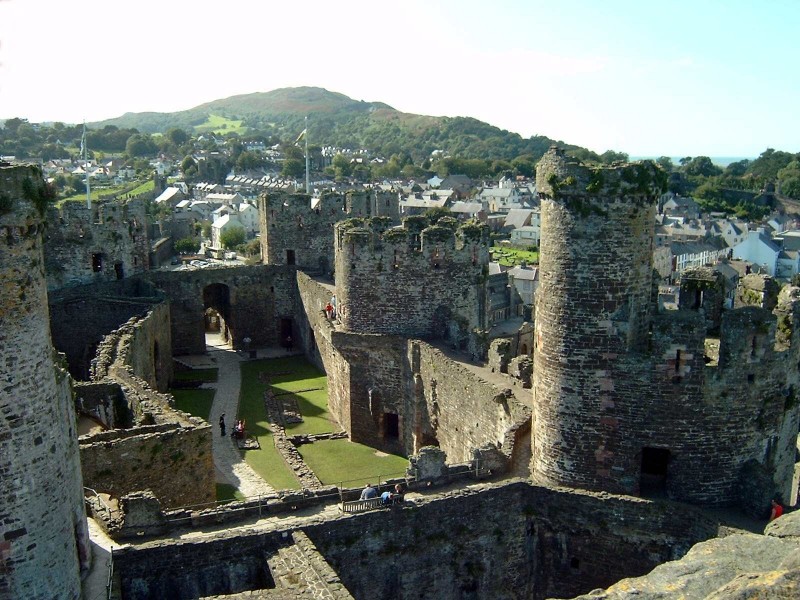}
        \end{minipage}
        \begin{minipage}{0.15\linewidth}
        \scriptsize{
	Who designed this building? \vspace{0.15cm}\\
        \textbf{LLaVA-1.5:}\\
        Architect \textcolor{red}{\xmark} \\
        \textbf{\ours:}\\
        James of Saint George \textcolor[HTML]{00b050}{\cmark}
        }
    \end{minipage}
        
    \vspace{0.15cm}

    \begin{minipage}{0.165\linewidth}
        \includegraphics[width=0.97\linewidth]{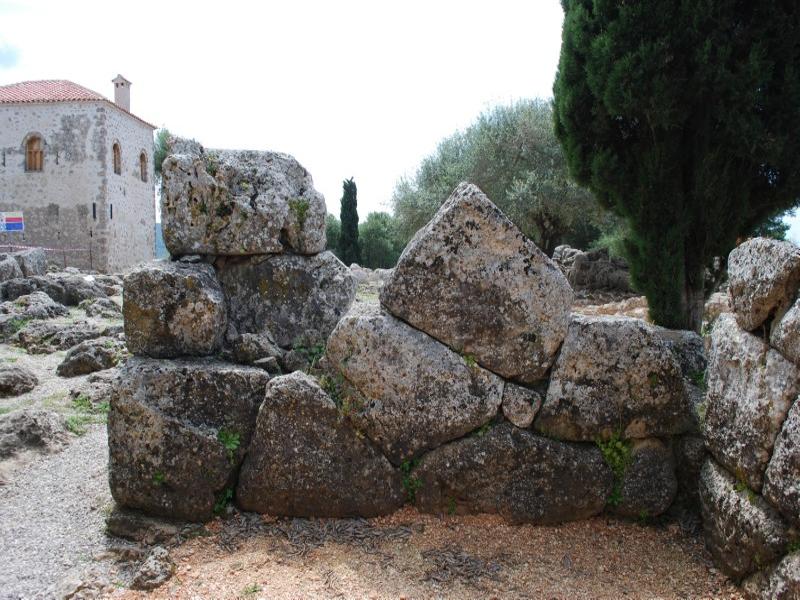}
        \end{minipage}
        \begin{minipage}{0.16\linewidth}
        \scriptsize{
	Which culture is associated with this place?\\ \texttt{Ancient Greek} \vspace{0.15cm}\\
        \textbf{LLaVA-1.5:}\\
        Roman \textcolor{red}{\xmark} \\
        \textbf{\ours:}\\
        Nuragic Civilization \textcolor{red}{\xmark}
        }
    \end{minipage}
    \hspace{0.02cm}
        \begin{minipage}{0.165\linewidth}
        \includegraphics[width=0.97\linewidth]{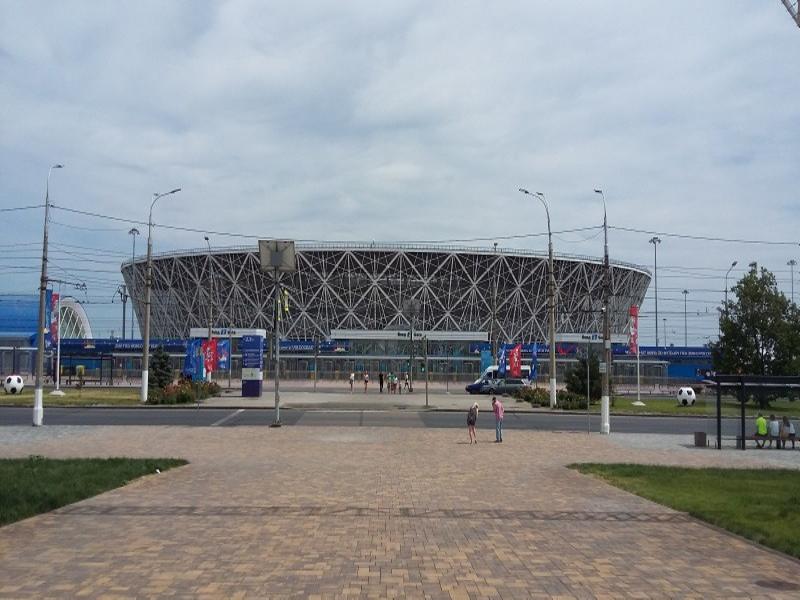}
        \end{minipage}
        \begin{minipage}{0.15\linewidth}
        \scriptsize{
	What is the name of the main club of this stadium? \\ 
    \texttt{FC Rotor}\vspace{0.15cm}\\
        \textbf{LLaVA-1.5:}\\
        Real Madrid \textcolor{red}{\xmark} \\
        \textbf{\ours:}\\
        FC Dynamo Kyiv \textcolor{red}{\xmark}}
    \end{minipage}
    \hspace{0.02cm}
        \begin{minipage}{0.165\linewidth}
        \includegraphics[width=0.97\linewidth]{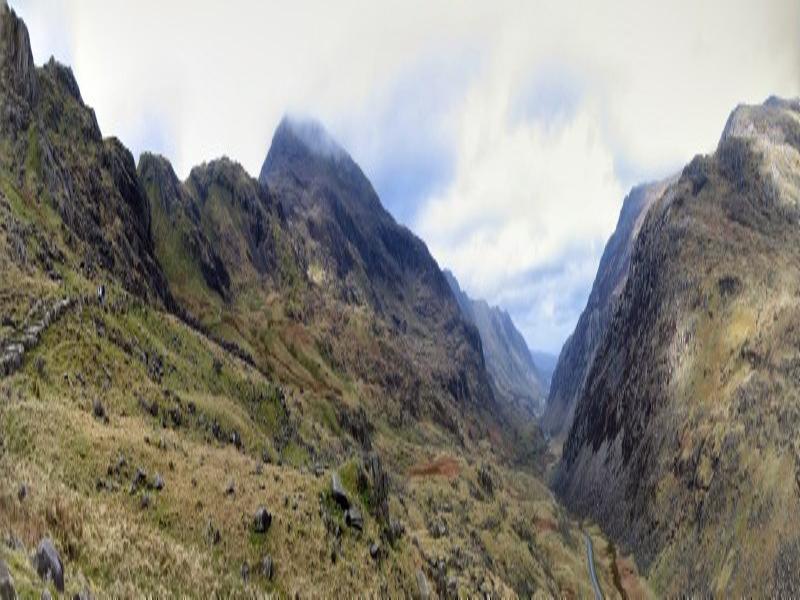}
        \end{minipage}
        \begin{minipage}{0.15\linewidth}
        \scriptsize{
	Which mountain range is this mountain belong to?\\ \texttt{Snowdonia}\vspace{0.15cm}\\
        \textbf{LLaVA-1.5:}\\
        Rocky mountains \textcolor{red}{\xmark} \\
        \textbf{\ours:}\\
        Lake District \textcolor{red}{\xmark}
        }
    \end{minipage}
    \vspace{-0.1cm}
    \caption{Qualitative results on sample image-question pairs from Encyclopedic-VQA (first row) and InfoSeek (second row) comparing the proposed approach with the original LLaVA-1.5 model. Some failure cases are shown in the third row with the corresponding ground-truth.}
    \label{fig:qualitatives}
    \vspace{-0.2cm}
\end{figure*}

Some qualitative results on sample image-question pairs from Encyclopedic-VQA (first row) and InfoSeek (second row) are reported in Fig.~\ref{fig:qualitatives}, comparing the answers given by \ours with those coming from the original LLaVA-1.5 model. For completeness, we also report some failure cases (third row) in which both models are not able to correctly answer the given question.

\tit{Evaluating the importance of the fine-tuning datasets}
As described in Sec.~\ref{sec:training} and Sec.~\ref{sec:details}, the MLLM fine-tuning is done with a mixture of data containing image-question-answer triples from the Encyclopedic-VQA or InfoSeek training set and visual instruction tuning data from LLaVA-Instruct~\cite{liu2023visual}, which has been used to originally fine-tune the LLaVA model. In Table~\ref{tab:ablation}, we evaluate the effect of mixing fine-tuning data for the knowledge-based VQA task. In this setting, we only report the results of the fine-tuned models without external knowledge retrieval. Notably, using visual instruction tuning data can help to regularize the fine-tuning phase on the InfoSeek dataset, leading to an overall improvement of 1.9 accuracy points compared to the model fine-tuned only on image-question-answer triplets from the training set of the dataset. On Encyclopedic-VQA, instead, training with instruction tuning data does not lead to performance improvement although without degrading the original results.

\begin{table}[t]
    \centering
    \setlength{\tabcolsep}{.25em}
    \resizebox{\linewidth}{!}{
    \begin{tabular}{lc cc c c c c c cc}
    \toprule
         & & \multicolumn{2}{c}{\textbf{MME}} & & \multicolumn{1}{c}{\textbf{MMMU}} & & \multicolumn{1}{c}{\textbf{MMB}} & & \multicolumn{2}{c}{\textbf{POPE}} \\
         \cmidrule{3-4} \cmidrule{6-6} \cmidrule{8-8} \cmidrule{10-11}
         \textbf{Fine-tuning} & & Cogn & Perc & & Acc & & Acc & & Acc & F1 \\
         \midrule
         - & & 355.7	& 1513.3 & & 35.1 & & 71.6 & & 86.9 & 85.8\\
         \midrule
         Enc-VQA & & 200.7 & 802.8 & & 36.6 & & 67.7 & & 72.9 & 63.4 \\
         Enc-VQA + LLaVA-Instruct & & 290.0 & 1170.1 & & 36.6 & & 70.4 & & 87.2 & 86.6 \\
         \midrule
         InfoSeek & & 296.8 & 1377.2 & & 35.2 & & 71.7 & & 82.0 & 79.6 \\
         InfoSeek + LLaVA-Instruct  & & 341.3 & 1438.9 & & 35.6 & & 71.1 & & 85.8 & 84.2 \\
    \bottomrule
    \end{tabular}
    }
    \vspace{-.1cm}
    \caption{Performance preservation analysis with respect to the original LLaVA-1.5 model (first row) on diverse benchmarks for MLLM evaluation.}
    \label{tab:llava_eval}
      \vspace{-0.35cm}
\end{table}

\tit{Preservation of LLaVA performance}
Finally, we analyze the impact of LLaVA fine-tuning on knowledge-based VQA datasets when evaluating the model on common MLLM evaluation benchmarks~\cite{caffagni2024r}. In particular, we include results on MME~\cite{fu2023mme} which contains image-question pairs covering 14 different tasks grouped in two macro-categories (\ie, cognition and perception), MMMU~\cite{yue2023mmmu} that is composed of multiple-choice and open-ended questions possibly interleaved with one or more images and extracted from diverse university textbooks and online courses, MMBench (MMB)~\cite{liu2023mmbench} that includes multiple-choice questions across 20 different domains, and POPE~\cite{li2023evaluating} that is focused on evaluating object hallucinations and comprises binary classification entries, each related to an image. More details about the evaluation metrics and number of samples can be found in the original paper of each dataset. 

Results are shown in Table~\ref{tab:llava_eval} comparing the original LLaVA model with the two fine-tuned versions on Encyclopedic-VQA and InfoSeek, with and without the use of visual instruction tuning data. Overall, employing samples from the LLaVA-Instruct dataset can better preserve the results of the original model, only partially degrading the performance on the considered benchmarks compared to the original model. While the most significant deterioration is achieved on the MME dataset, in the other settings the original performances are better preserved, also leading to a slight improvement on MMMU and POPE benchmarks compared to the LLaVA-1.5 results.

\subsection{Limitations and Future Works}
While our work provides an initial step towards MLLM which can properly exploit external multimodal data, it is worthwhile mentioning that significant research is needed in two directions. The fist is defining proper embedding spaces in which documents can be retrieved from questions and input images, so as to improve the performance of the higher level of our hierarchical retrieval. The second is modeling an efficient and sustainable paradigm to select from one or more documents. Here, the challenge is to 
increase the capability of the MLLM of distinguishing the appropriateness of retrieved items. This point might also require novel architectural design, which might go beyond the pure inclusion of retrieved items in the context.
Regardless of its current limitations, our research testifies the potential of adding multimodal external knowledge to a MLLM and inherits all the advantages of retrieval-augmented approaches, such as the adaptability to different domains and the loosely-coupled relationship between pre-trained information and retrievable data.    
\section{Conclusion}
\label{sec:conclusion}
We have presented \ours, an architecture for augmenting an existing MLLM with external knowledge. Our proposal leverages an external knowledge source of documents to improve the effectiveness of an MLLM when tasked with questions and dialogues. In particular, we devise a hierarchical architecture for retrieving documents and eliciting selected parts to be included in the MLLM input context. Extensive experiments demonstrate the effectiveness of the proposed solution, and its capability to maintain the proficiency of the MLLM across different tasks.

\section*{Acknowledgments}
We acknowledge the CINECA award under the ISCRA initiative, for the availability of high-performance computing resources and support. This work has been conducted under two research grants, one co-funded by Leonardo S.p.A. and the other co-funded by Altilia s.r.l., and supported by the PNRRM4C2 project ``FAIR - Future Artificial Intelligence Research'', funded by the European Commission, and by the PNRR project ``Italian Strengthening of Esfri RI Resilience'' (ITSERR) funded by the European Union - NextGenerationEU (CUP B53C22001770006).

{
    \small
    \bibliographystyle{ieeenat_fullname}
    \bibliography{main}
}

\end{document}